\documentclass{acm_proc_article-sp}
\usepackage{url}
\usepackage{graphicx} 
\usepackage{colortbl}
\usepackage{epstopdf}
\begin{document}

\title{Inferring land use from mobile phone activity\titlenote{Permission to make digital or hard copies of all or part of this work for personal or classroom use is granted without fee provided that copies are not made or distributed for profit or commercial advantage and that copies bear this notice and the full citation on the first page. To copy otherwise, or republish, to post on servers or to redistribute to lists, requires prior specific permission and/or a fee.
UrbComp'12, August 12, 2012. Beijing, China. 
Copyright 2012 ACM 978-1-4503-1542-5/08/2012 É\$15.00.}} 
%
%
%
%
%

\numberofauthors{4} 
%
\author{
%
%
\alignauthor
Jameson L. Toole\titlenote{Please direct all correspondence to Jameson Toole at \email{jltoole@mit.edu}}\\
       \affaddr{Massachusetts Institute of Technology}\\
       \affaddr{77 Mass. Ave}\\
       \affaddr{Cambridge, MA, USA}\\
       \email{jltoole@mit.edu}
\alignauthor
Michael Ulm\\
       \affaddr{Austrian Institute of Technology}\\
       \affaddr{Vienna, Austria}\\
       \email{michael.ulm@ait.ac.at}
\alignauthor Marta C. Gonz\'alez \\
       \affaddr{Massachusetts Institute of Technology}\\
       \affaddr{77 Mass. Ave}\\
       \affaddr{Cambridge, MA, USA}\\
       \email{martag@mit.edu}
\and  
\alignauthor
Dietmar Bauer\\
       \affaddr{Austrian Institute of Technology}\\
       \affaddr{Vienna, Austria}\\
        \email{dietmar.bauer@ait.ac.at}
}
\date{\today}

\maketitle
\begin{abstract}
Understanding the spatiotemporal distribution of people within a city is crucial to many planning applications. Obtaining data to create required knowledge, currently involves costly survey methods.  At the same time ubiquitous mobile sensors from personal GPS devices to mobile phones are collecting massive amounts of data on urban systems. The locations, communications, and activities of millions of people are recorded and stored by new information technologies. This work utilizes novel dynamic data, generated by mobile phone users, 
to measure spatiotemporal changes in population.  In the process, we identify the relationship between land use and dynamic population over the course of a typical week.  A machine learning classification algorithm is used to identify clusters of locations with similar zoned uses and mobile phone activity patterns.  It is shown that the mobile phone data is capable of delivering useful information on actual land use that supplements zoning regulations. 
\end{abstract}



\begin{figure}[h] 
\centering
\includegraphics[width=1\linewidth]{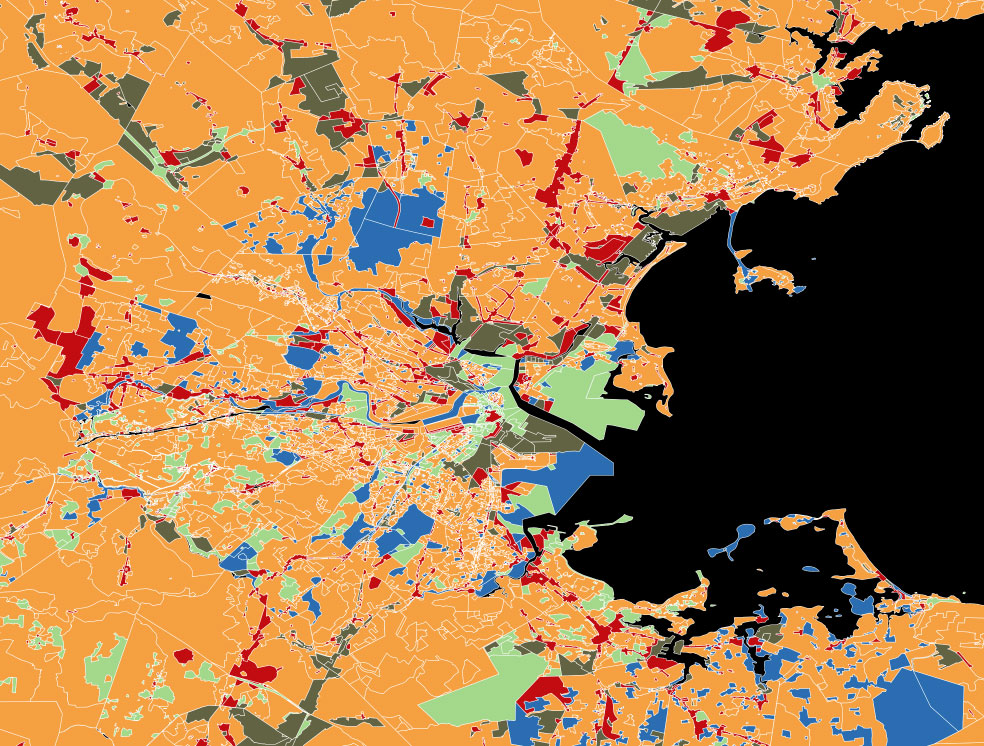}
\caption{Zoning regulation for the Boston area. Color code: orange - Residential, red - Commercial, gray - Industrial, blue - Parks, green - Other.}\label{fig:zone_bos} 
\end{figure}

\section{Introduction}
In describing the ``organized complexity" of cities, Jane Jacobs notes that a "park's use depends, in turn, on who is around to use the park and when, and this in turn depends on uses of the city outside the park itself." \cite{Jacobs1961} Where people live, work, and play is intimately related to the time and distance required to move to and from these locations \cite{Geurs2004}.  Understanding how individuals are distributed in space and time is crucial to making effective and efficient planning decisions within cities. For example, the location choices of residents and firms is influenced by and determines the demand for mobility.  Restaurants  want to maximize patronage by choosing a popular location and individuals want to maximize their access to amenities.
 
How a particular area of a city is used is determined, in part, by the zoning regulations implemented and enforced by local governments. These regulations impact the structure of a city by dictating where housing or office space can be located.  Zones of a kind share common usage. The central business district (CBD) for instance is populated during office opening hours whereas when offices are closed, relatively few people are found in these zones. Different zones relate to different land use which is related to different population size to be found at any given time in the zone. In practice, however, many zones feature different usage which might also differ somewhat from intended use. As an example zoning information for the Boston area is shown in Figure~\ref{fig:zone_bos}. Note, that zoning areas are not only restricted to land but also cover parts of rivers, lakes and the sea.

There is a large body of work dedicated to understanding the spatiotemporal dynamics of population and its relation to land use \cite{Maat2005, Banister1997, Cervero1996}.  Measurements of human mobility within cities has traditionally been made via travel surveys. These surveys require subjects to record data on where they are moving to and from in the observation period (typically one day or a whole week), how they are doing so, and why.  However, because surveys typically feature in-person interviews and demand a high workload for each subject, this method of data collection is expensive and limited. 

Given these limitations, travel surveys suffer from relatively small samples (usually below tens of thousands of individuals), capture only short periods for each individual, and are updated 
infrequently.  Fortunately, over the past decade a new type of measurement instrument has made its way into the pockets of people in nearly every culture and country.  Each of the roughly 6 billion mobile phones currently in use \footnote{\url{http://www.itu.int/net/ITU-D/index.aspx}} is capable of recording the location of calls, SMS, and data transmissions to within a few hundred meters.  Moreover, these data are also collected centrally by mobile phone providers for billing purposes. With these data come enormous opportunities to improve our understanding of human mobility patterns.

In particular, call detail records (CDR) data, which provide information on the location of mobile phones any time a call is made or a text message is sent, contain much information on the distribution of persons in a region.  This information can be obtained at low costs.  Moreover, aggregated data only contains the number of active phones in a given area during a given time interval. This method of data collection provides much higher levels of anonymity reduces the risk any breach of individual information.  Given the (imperfect) relation between the distribution of persons and active phones in a region the question arises as to whether the distribution of the numbers of active mobile phones can be used in order to infer land usage in a given zone.

To have such a measurement method would be very advantageous. Corresponding results can be used to monitor the use of all zones of a given zone class.  Zoning regulation that all zones of one class share a common usage whereas the usage might differ for a number of reasons. Knowledge on different usage can be used to understand demand for mobility infrastructure across space and time. Monitoring the usage over time allows to detect changes in habits of the population as well as shifts in usage which may indicate ongoing regional developments. 

Consequently this work investigates the potential of applying aggregated CDR data in order to infer dynamic land use, i.e. to understand how the population of different areas of a city changes with time and according to specific zoned land uses. The work centers on supervised classification of regions according to given zoning regulations. We demonstrate that CDR data can be used in order to classify zones of different types with reasonable accuracy. To this end, normalization techniques are discussed to highlight differences between zones. Then, the application and result of random forests for the classification is described in detail.

\section{Mobile phones and human mobility}
Mobile phones have proven good instruments to measure human behavior.  In one of the first studies utilizing these devices, Eagle and Pentland~\cite{Eagle2006} were able to decompose mobile phone activity patterns of university students and employees into regular daily routines. Moreover, these patterns were found to be predictive of an individual's characteristics such as their major or employment level (i.e. graduate student).  Subsequent research has built upon this work, scaling up in both geographic extent and sample size.  Gonz\'alez et al \cite{Gonzalez2008} studied data from nearly one-hundred thousand anonymous mobile phone users to reveal persistent regularities in the statistical properties of human mobility.  Highlighting the remarkable predictability of human behavior, Song et al \cite{Barabasi2010} estimated that it is theoretically possible to predict individual movements of users with as high as 93\% accuracy using only data from mobile phones.

Mobile phone data has also been used to study how space is used over time.  Reades et al \cite{Reades2009} used mobile phone network data from Rome, Italy, to link mobile phone activity to commercial land uses. Measuring mobile phone activity in 1km by 1km grid cells, they employ a form of principal component analysis to identify the dominant activity patterns.  
The authors qualitatively interpret areas of the city exhibiting this signal as Commercial, though actual zoning information is not introduced. They then decompose activity across the city to identify regions with similar patterns of usage.  Similarly, Soto et al \cite{Soto2011} use CDR mobile phone data at the cell tower level to identify clusters of locations with similar activity.  Qualitative agreement between these clusters and land uses were observed. 

Calebrese et al \cite{Calabrese2010} have applied similar decomposition and clustering techniques to classify locations on a university campus as classrooms, dormitories, etc. By analyzing wifi activity across 3000 wifi-access points, the authors used unsupervised, non-parametric techniques to identify clusters of similarly used locations.  These locations naturally fit into location profiles such as "lecture hall" or "dormitory."  Finally, CDR data have proven useful to detect movement at the census tract scale \cite{Calabrese2011}. Location data from calls helped to measure origins and destinations for trips across the Boston Metropolitan area.  However, no attempt was made to associate such trips with land uses.

Other data sources such as points of interest (POIs) as well as GPS data collected from taxi fleets have been combined with unsupervised learning algorithms to identify the rich structure of different functional sections of a Beijing \cite{Yuan2012}.  To date, however no studies exist that employ supervised learning techniques to combine traditional data sources on land use such as zoning regulations and CDR data. This study aims to investigate the link between zoned land use and mobile phone activity on a common spatial partitioning of the greater Boston area into regions of homogeneous land use.  For each region the temporal profile of active phones is used in supervised classification techniques in order to identify patterns characteristic for a specific zoning classification. The corresponding patterns will be interpreted in detail. 


\section{Data sources}
Two data sources are used in this work: mobile phone activity records and  zoning regulations. For the Boston metro region, anonymized CDR provide the location of a mobile phone by triangulating signal strengths from surrounding cell towers, unlike traditional CDR data, in which record the location of a call as the location of the mobile phone tower.  This provides slightly higher accuracy and allows us to measure calls continuously across space rather than at points where towers are located.  Triangulation by this method is accurate to within a few hundred meters depending on the tower density. These data make it possible to measure the amount of phone activity (counts of the number of calls and texts) that occurs within a given area and time window.  In this study we use three weeks of CDR data for roughly 600,000 users in the Boston region home to roughly 3 million people.  Though mobile phone data come from specific set of carriers, the market share of these carriers is between 30\% and 50\%.

In addition to mobile phone activity, we obtain zoning classifications for the Boston metropolitan area. The Massachusetts Office of Geographic Information (MassGIS) aggregates uses into five categories: Residential, Commercial, Industrial, Parks, and Other. We are careful to note our assumption that actual land use and zoning classification are closely related while acknowledging that zoning regulations are only a proxy of actual land use imposing restrictions.

\section{Common spatial representation}
The first obstacle to studying the relationship between phone activity and land use is the reconciliation of the spatial dimensions of the data: While the location of the phone activities are recorded as coordinate pairs, zoning data is provided in polygons at roughly the parcel scale.  The spatial partitioning of phone and population data is rarely the same as zoning parcels.  To reconcile all data sources as well as to reduce the influence of noise (due to inter alia sources localization estimation noise) in the data, we transform both to the same uniform grid.  A lattice is laid over the analysis region such that every cell in the lattice measures 200 by 200 meters. Different grid sizes have been tested, 200 meters proved to be a good aggregation level; being coarse enough to reduce the noise level and detailed enough in order not to mix many parcels of different zoning areas. 

In order to reduce the high noise level average hourly time series of phone activity are computed. Here, the average is computed for each hour within a day of the week.  Only cells with mobile phone activity above a certain threshold are used in the analysis.  

With respect to zoning data, each cell is given a single zoning classification based on the most prevalent (in terms of fraction of area covered) use within the area. 

\begin{figure}[h] 
\centering
\includegraphics[width=1\linewidth]{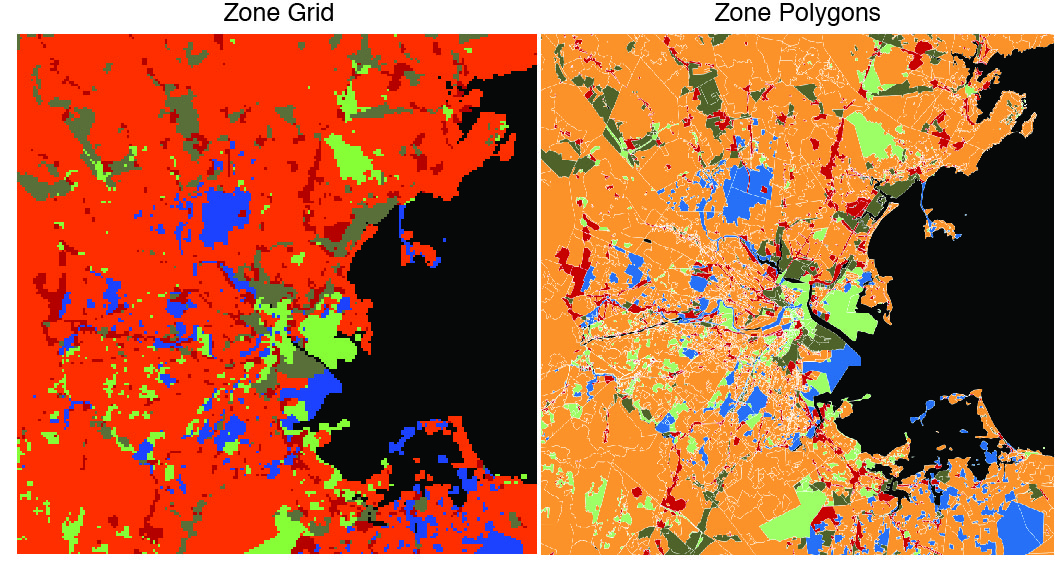}
\caption{To improve computational efficiency and reconcile all mobile phone and traditional data sources, we create a uniform grid over the city.  Zoning polygons (right), are rasterized to cells 200m by 200m in size (left).  For cells where more than one zoning class exists, the most prevalent class is used. Given the small size of these cells, this data transformation provides an accurate map of the city while improving computational efficiency.}\label{fig:zone_compare}
\end{figure}

Potential pitfalls of this method arise due to large heterogeneity in population density. Downtown areas are much more densely populated than the suburbs, a characteristic that is reflected in other spatial divisions like census tracts.  This leads to sparse mobile phone activity in rural regions.  However, the small grid size used in this analysis retains detailed information about block to block zoning regulations in dense urban areas.  Figure \ref{fig:zone_compare} displays actual zoned parcels versus the gridded approximations.

Table \ref{tab:bos_zones_tab} shows the frequency of each zoning class in the grid.  The vast majority of land, nearly 75\% of cells, are zoned as Residential. Other uses appear in roughly equal fractions.
\begin{table}[h] 
	\caption{Tabulation of Boston zoning.  The land use profile of the city is dominated by residential use accounting for nearly 75\%. Other uses share roughly the same percentage of remaining land.}
	\label{tab:bos_zones_tab}
	
		\begin{tabular}{cccc} 
		\hline
			Zone Use & Category Index & Count & Percentage \\ 
			\hline
		Residential &  1  &  23322 & 74.28\\
		Commercial &  2 & 1854 &  5.90\\
		Industrial &  3 & 2236 &  7.12\\
		Parks &  4 & 1941 &  6.18\\
		Other &  5 & 2045 &  6.51\\
		\hline
		\end{tabular}
	
\end{table}

\section{Descriptive Statistics}
\begin{figure*}[ht] 
\centering
\includegraphics[width=1\linewidth]{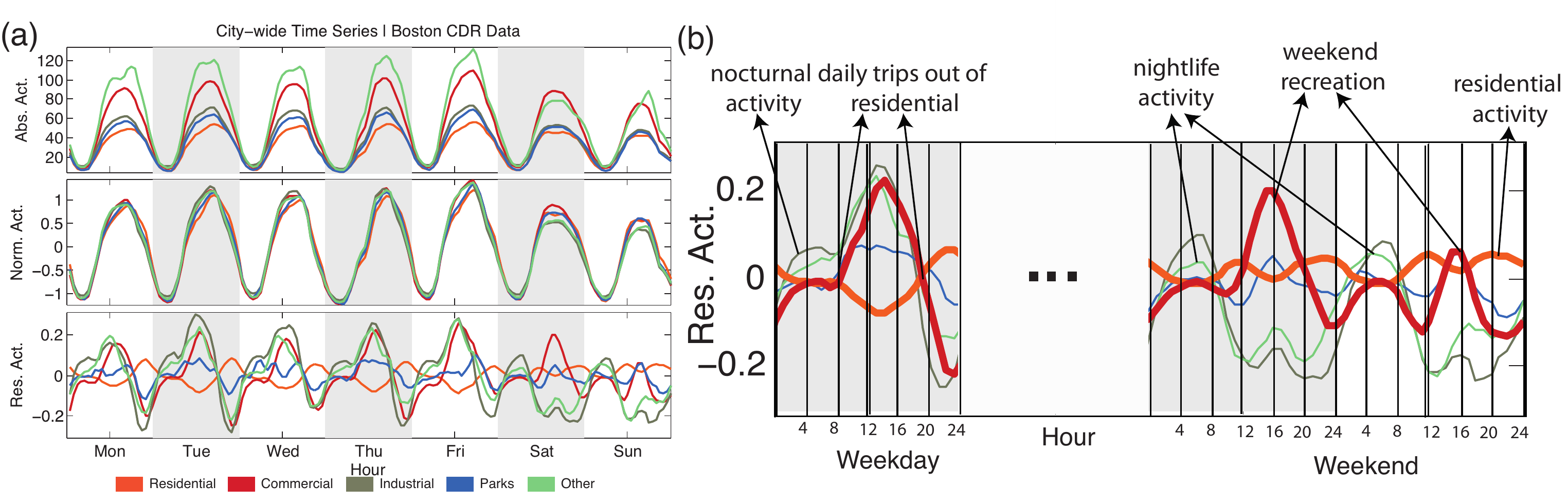}
\caption{(a) Plots are shown for three different time series of average mobile phone activity within each of five land use.  The first plot shows absolute activity (number of calls and SMS messages).  The second plot displays z-scored time series. 
The bottom plot shows residual activity. 
(b) More detailed view of average (over cells of the same zoning class) residual activity. 
}\label{fig:macro_ts}
\end{figure*}

We first examine the relationship between mobile phone activity and land use at the macro, city-wide scale.  Figure \ref{fig:macro_ts} displays time series of mobile phone activity averaged over all cells of a given zoning classification.  Examining absolute counts (first row) reveals that the average activity level in different zoning classifications differs greatly. While residential areas only show a maximum activity of roughly 50 events per hour, commercial cells reach approximately 100 events on average.  

The number of activities within different cells shows huge differences.  The downtown area of Boston shows orders of magnitude higher activity levels than typical residential zones. In order to allow for classification based on relative mobile phone activity, time series are normalized using a z-score. By definition, the normalized time series have zero mean and unit standard deviation. Mathematically, the normalized activity of cell ($i$,$j$) is given by:
\begin{equation}
\label{eqn:res_act}
a_{ij}^{norm}(t) = \frac{a_{ij}^{abs}(t) - \mu_{a_{ij}^{abs}}}{\sigma_{a_{ij}^{abs}}}
\end{equation}
The second row of Figure \ref{fig:macro_ts} (a) shows the average (over cells of one zoning class) normalized activity. These profiles are remarkably similar for all zoning classes showing the strong circadian rhythm of the city.  Residents wake up, go to sleep, and wake again the next day.  The rise and fall of activity in each zone, however, is not solely the result of users moving into and out of a region, but is instead also partly due to an uneven distribution of phone use across the day. To account for this, during each hour, we subtract the average normalized activity of the entire region from the normalized activity at each given cell. The corresponding spatially de-meaned series will be referred to as {\em residual activity}. Residual activity can be interpreted as the amount of mobile phone activity in a region, at a given time, relative to the expected mobile phone activity in the whole city at that hour.  Mathematically, it is calculated as follows:
\begin{equation}
a_{ij}^{res}(t) =  a_{ij}^{norm}(t) - \bar{a}^{norm} (t)
\end{equation}
where $\bar{a}^{norm} (t)$ is the normalized activity averaged over all cells at each particular time.
Averaging the residual activity for each zoning classification reveals patterns related to travel behavior. 
The last row of Figure \ref{fig:macro_ts} (a) and (b) provide the residual activity averages across zoning classes for weekdays and weekends. The most notable signal is the inverse relationship between residual activity in residential and commercial areas: While residential areas on average show higher than expected activity during the night and lower than expected during weekdays. As expected, the opposite is true for commercial zones. Somewhat surprisingly, the normalized activity does not show these features strongly. Only the residual activity demonstrates the expected behavior. There, also higher than average activity in parks on the weekend afternoons is visible.
 
Residential areas have higher residual activity in the early morning hours and late at night, while commercially zoned cells have a peak period during the day and show much lower activity levels late at night. These patterns most likely reflect the 9-to-5 business hours of offices and stores. More subtle patterns are also visible. In Boston, much of the CBD is zoned as Other or Mixed use. We see that residual phone activity in this zoning type has peaks in the early morning hours on Saturday and Sunday, suggesting these areas support night life on the weekends. These city-wide time series show that mobile phone activity and land use are linked at the highest level of aggregation.  By treating phone activity as a proxy for the spatial distribution of people at a given time period the expected patterns of concentration of people in the CBD and inner city region during the working day, and the shifts induced by the commuting behavior are visible in the residual activity levels.

We note that because residual activity is relative to absolute call volume as well time of day, it is not affected by differences in mobile phone usage across zoned uses provided those differences are persistent in time.  For example, it does not affect measurements if individuals are twice as likely to make a phone call in a commercial zone than a residential zone as long as this propensity is constant across all hours of a week.

\begin{figure}[ht] 
\centering
\includegraphics[width=1\linewidth]{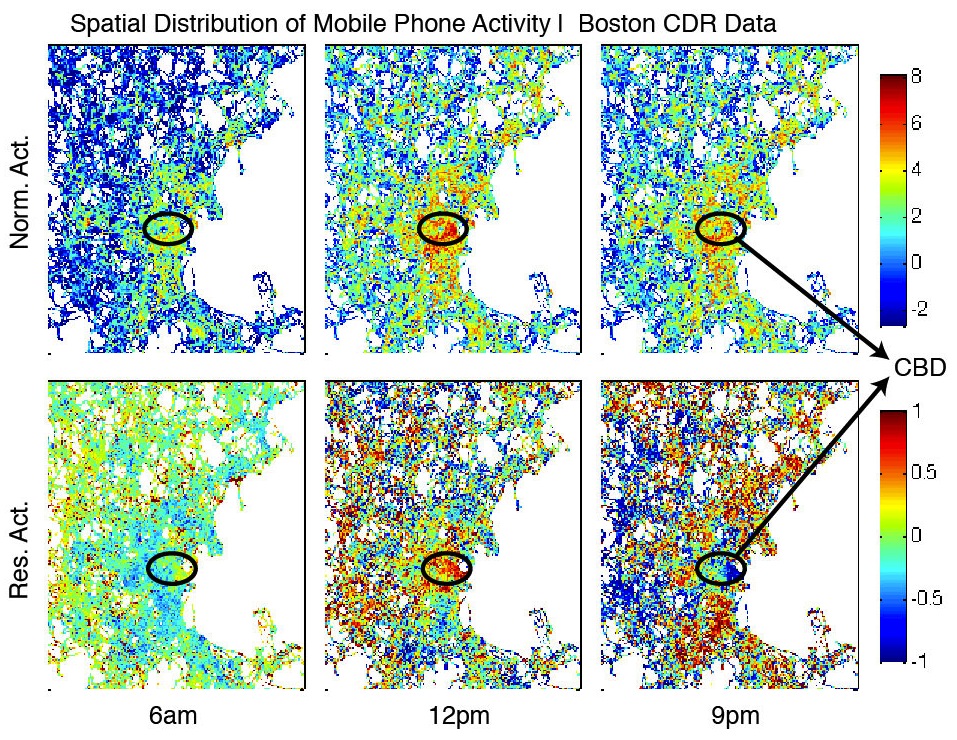}
\caption{Spatial distribution of absolute and residual phone activity over the course of a day.  While absolute mobile phone activity is dominated by population density and sleep and wake patterns, residual activity reveals flows into and out of the city center over the course of a day.}\label{fig:macro_space}
\end{figure}

Figure \ref{fig:macro_space} displays the spatial distribution of normalized activity (top row) and residual activity (bottom row) at three time instants. Not shown in the plots are the absolute activity levels which are distributed much like population density. The CBD of Boston has orders of magnitude more activity than the rest of the city.  Mapping the logarithm of absolute activity over time once again only reveals the circadian rhythm of the city which strongly dominates the differences in land usage which consequently are not seen in these plots.

In the spatial distribution of the normalized activity the dominance of the CBD is less pronounced. Nevertheless, the circadian rhythm still dominates the differences between different zones. From this perspective, Boston appears as a monocentric region, with small pockets of density located on an urban ring roughly 20km from the CBD.  

By way of contrast, the spatial distribution of residual activity reveals a much richer structure.  
In the early morning hours, residual activity is located on the periphery of the region.  During the day, this activity becomes heavily concentrated in the CBD or in small subcenters on the urban 
ring. Later in the evening, activity again returns to residential areas on the periphery, away from centers. This suggests some correlation between commuting patterns and the spatial distribution of residual activity. 
\section{Classifying Land Use by Mobile Phone Activity}
In the last section we observed correlation between residual mobile phone activity and land use on the macro scale.  Fluctuations in mobile phone activity mimics intuition of population changes related to commuting and recreational trips.  In this section we investigate the question whether usage of cells of one zone class are homogeneous. This will be done by performing supervised classification based on features extracted from the residual activity time series and the classes provided by the zoning regulations as labels.  Though previous work in this area has employed unsupervised learning techniques, access to extensive zoning data in a mature, regulated city such as Boston makes supervised learning an attractive option.  Cross validation is used to test performance.    


We implement the {\it random forest} approach described by Breiman \cite{Breiman2001}. 
Other approaches including neural network based classifiers have been tested and led to similar results. Random forests are useful for their ability to efficiently classify data with large numbers of input variables (such as long time series).  Rather than make comparisons for every feature of the data every time, a number of random subsets are chosen to more efficiently search the space.  This does not come at the cost of accuracy as random forests have been shown to have high performance on a variety of datasets~\cite{Breiman2001}.  Moreover, random forest classifiers allow weights to be introduced so that more frequently occurring classes do not overwhelm smaller ones.  This feature will be exploited later to control for the large share of residentially zoned locations.


\begin{figure}[ht] 
\centering
\includegraphics[width=1\linewidth]{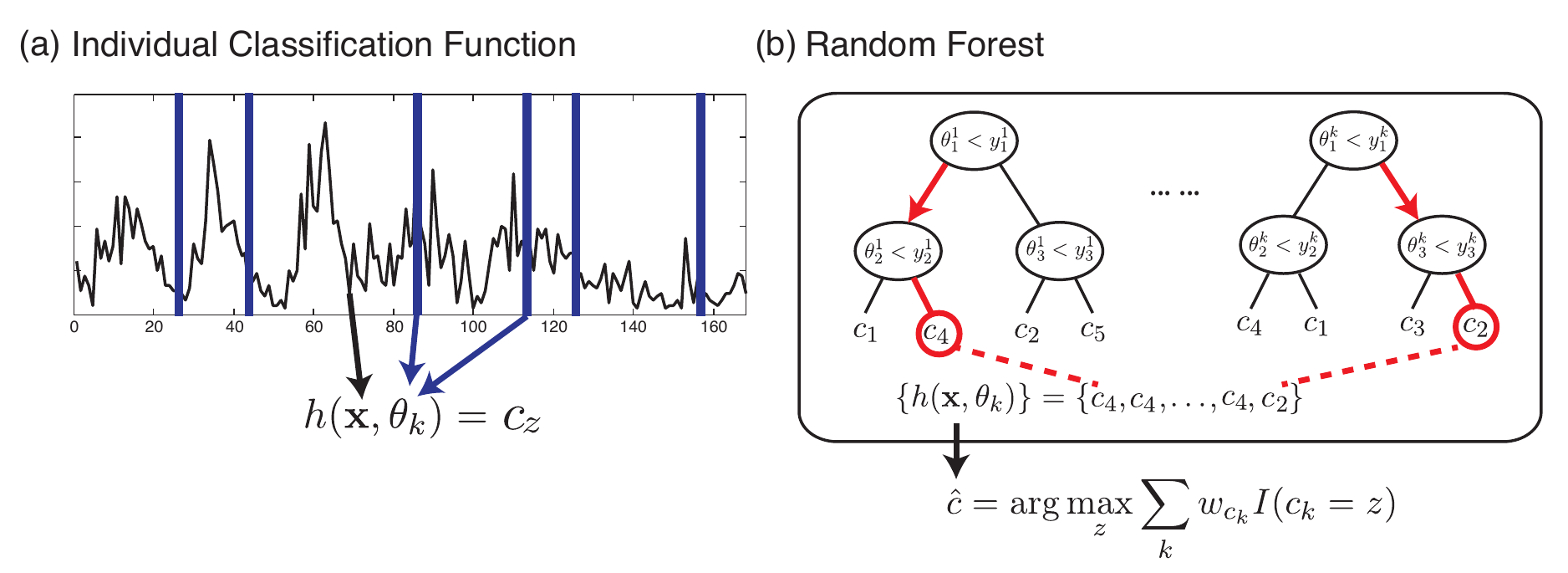}
\caption{(a) Shows the inputs to each decision tree $h(\mathbf{x},\theta_k$).  A time series of residual phone activity, $\mathbf{x}$, is input and activity at a random subset of times , $\theta_k$ (denoted by the blue bars), is chosen to make comparisons.  (b) A depiction of the random forest shows a number of different trees making predictions based on a different set of random times. Each tree casts a weighted vote for a certain classification. A final classification, $\hat{c}$, is made by counting these votes.}\label{fig:rf_desc}
\end{figure}

A random forest, $\{h(\mathbf{x}; \theta_k), k=1,...\}$, is constructed from a set of decision trees as visualized in Fig.~\ref{fig:rf_desc}. The training data is used to determine the parameter vectors $\theta^k$.  Least squares or maximum likelihood estimation can be used to find these configurations. To obtain a single prediction for each input time series, a voting scheme is implemented. Each tree votes for a class based on its prediction. These votes can be weighted (weights denoted by $w_{c_k}$) so that votes for one class count more or less than votes for a different class. The weighted votes are summed and a single zoning class prediction, $\hat{c}$ is chosen for the original input time series.

For the calculations we use a MATLAB implementation of the random forest algorithm released by Jaiantilal \footnote{\url{http://code.google.com/p/randomforest-matlab/}}. 
Our implementation uses 49 input features which are computed for each location as the input feature vector $\mathbf{x}$.  These features include a 24-hour time series of residual mobile phone activity during an average weekday as well as a 24-hour time series of residual activity for an average weekend-day. The final feature is the mean of the location's absolute activity on any given day.  Additional features such as the variance of mobile phone activity were tested, but none aided prediction.  The output of the algorithm is a zoning classification for each location. Cross validation is used to test accuracy.  We create 500 trees for each forest and define total accuracy as the fraction of correctly classified cells on the validation part of the sample.

Our first set of results include all five zoning classifications: Residential, Commercial, Industrial, Parks, Other. When all land use classes are included, however, we face a major challenge with classification.  As noted above, nearly 75\% of all cells are primarily residential.  The next most common zoned use is Industrial at 7\%. Because of our definition of total accuracy, the most naive classifier, simply assigning Residential to everything, will achieve 75\% total accuracy, but will fail to capture any diversity in use. To guard against this, we weight the voting system so raise or lower the required votes in order to choose a given classification.  The maximum of the weighted votes then provides the predicted class. Systematic variations of the weights on a (coarse) grid led to a choice of weights where the criterion applied was maximum classification accuracy for all classes but residential. 

Finally, we note that the random forest classifier uses local information only to make a prediction.  Given the size of our grid cells, it is reasonable to assume that land use does not differ greatly from each 200m by 200m tract of land to the next.  To incorporate neighborhood information into our predictions, we implement a second pass algorithm.  After the classifier has made a prediction for a cell, we examine the predictions for each of that cell's neighbors.  If the majority of neighboring cells were predicted to be a land use that differs from the cell in question, that cell is switched to the majority use of its neighbors.  In practice, this results in some spatial smoothing of noisy classification data.  We find that performing the second pass provides gains of 2-10\% overall accuracy for each classifier.

\begin{figure}[ht] 
\centering
\includegraphics[width=1\linewidth]{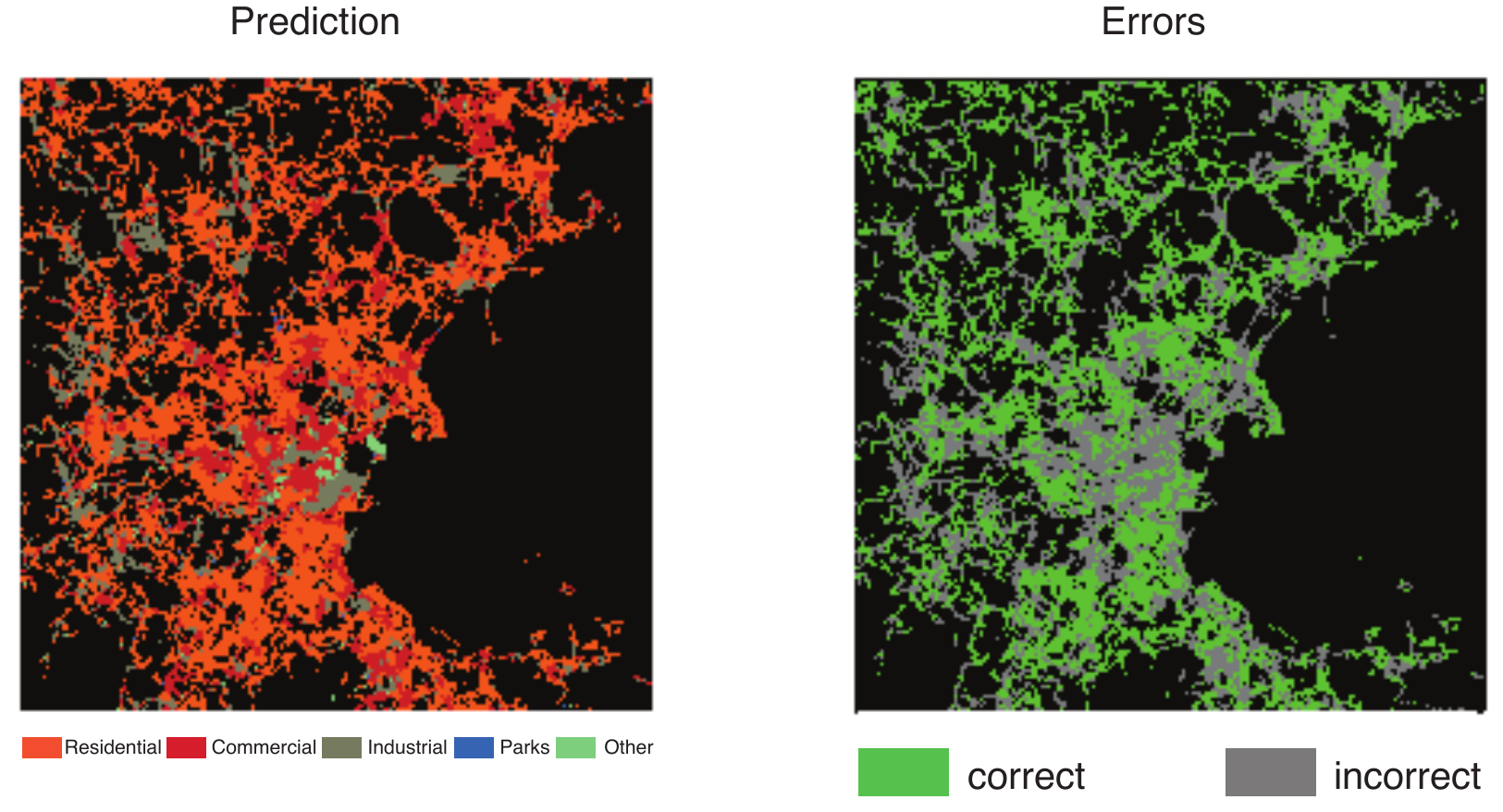}
\caption{Left plot: zoning map as predicted from mobile phone data using the random forest classification algorithm.  Right plot: spatial distribution of where the algorithm predicts land use correctly and where it fails. In general, these errors seem randomly distributed in space, suggesting that errors are not the result of some spatial correlations such as population density. For comparison to actual zoning, see the left panel of Figure \ref{fig:zone_compare}.}\label{fig:bos_class_result_33}
\end{figure}

Even with vote weighting and the second pass algorithm, we achieve only modest results.  
Table \ref{tab:bos_class_result_all_33} shows 54\% accuracy over the whole city.  This implies that demanding equal classification accuracy for all classes reduces overall accuracy by about 20\%.  Figure \ref{fig:bos_class_result_33} displays the spatial distribution of correctly and incorrectly classified locations.  We note, however, that the algorithm does capture some spatial patterns in the data and that our intra-use accuracy is relatively high for Commercial and Industrial uses.  Parks and Other mixed uses remain difficult to classify.  

\begin{table}[h] 
	\caption{Random forest classification results. The threshold refers the total number of phone events required in each cell over period of data collected to be considered for classification. Total accuracy is defined as the fraction of correctly classified cells.  The share refers to the percentage of cells actually zoned for each class of use.  Element $(i,j)$ of the confusion can be interpreted as the fraction of actual zoned uses of class $i$ that were classified as use $j$ by the random forest. Thus the high percentages in the Res column can be interpreted as the algorithm heavily favoring classification as residential due to its overwhelming share of overall uses.}
	\label{tab:bos_class_result_all_33}
	
\begin{tabular}{cccccc}
\hline 
 Total Accuracy: & \multicolumn{5}{l}{ 0.54} \\
\hline
 & Res & Com & Ind & Prk & Oth   \\
 Land Share:& 0.74 & 0.09 & 0.08 & 0.04 & 0.05\\
 Vote Thresh: & 0.60 & 0.10 & 0.10 & 0.10 & 0.10 \\
\hline
& \multicolumn{5}{c}{Confusion Matrix} \\
& Res & Com & Ind &Prk & Oth \\
Res & \cellcolor[gray]{0.8}0.62 & 0.21 & 0.15 & 0.01 & 0.01 \\
Com & 0.30 & \cellcolor[gray]{0.8}0.48 & 0.19 & 0.00 & 0.02 \\
Ind  & 0.33 & 0.27 & \cellcolor[gray]{0.8}0.38 & 0.00 & 0.02 \\
Prk &0.52 & 0.26 & 0.18 & \cellcolor[gray]{0.8}0.02 & 0.02 \\
Oth & 0.37 & 0.28 & 0.25 & 0.00 & \cellcolor[gray]{0.8}0.10\\
\hline
		\end{tabular}
	
\end{table}

\begin{figure}[h] 
\centering
\includegraphics[width=1\linewidth]{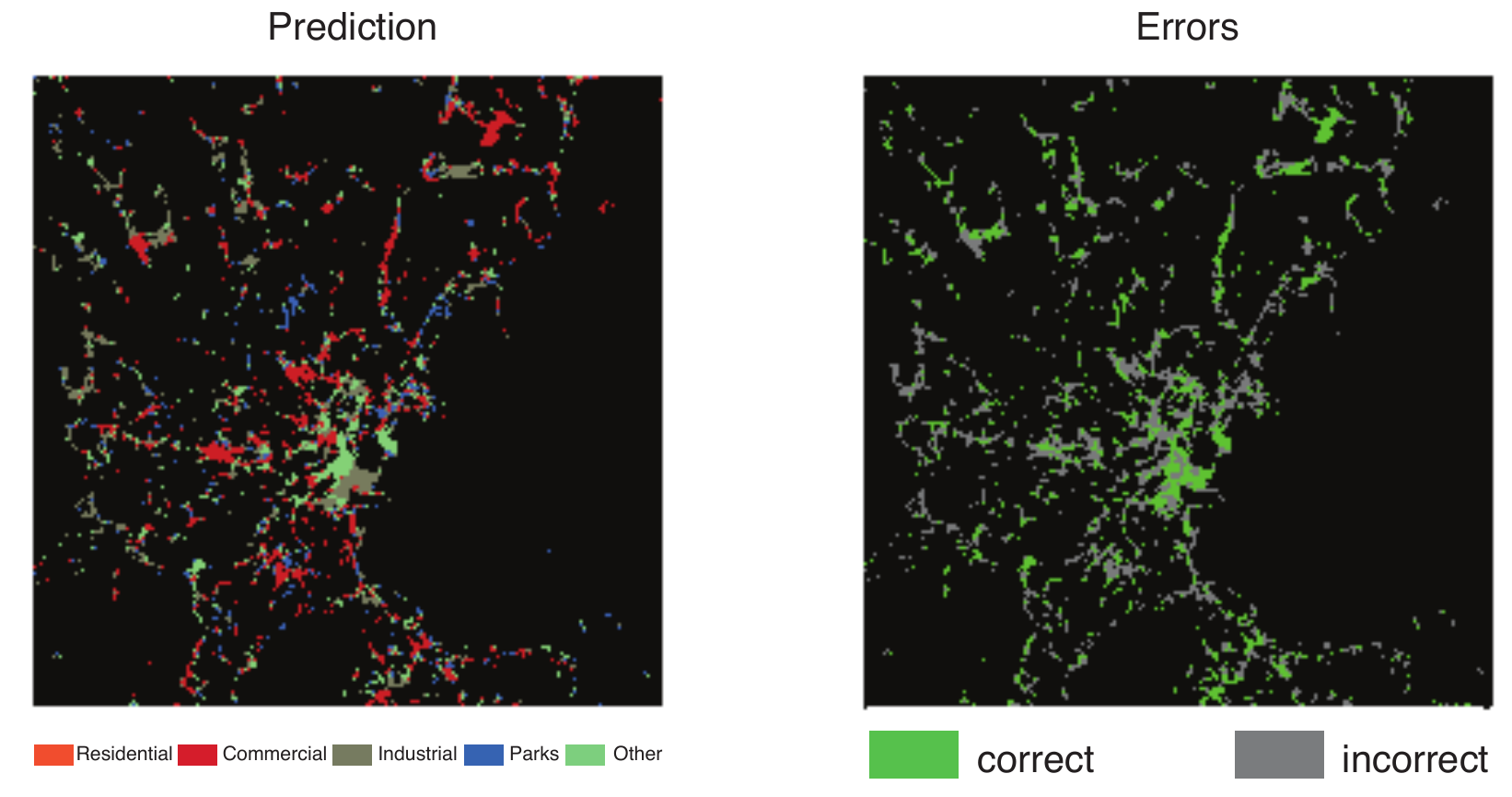}
\caption{The left plot shows the city zoning map with residential areas removed as predicted from mobile phone data using the random forest classification algorithm.  The right map displays the spatial distribution of where the algorithm predicts land use correctly and where it fails. Without residential areas to predict, the algorithm performs significantly better at predicting other uses. For comparison to actual zoning, see the left panel of Figure \ref{fig:zone_compare}.}\label{fig:bos_class_result_22}
\end{figure}

To account for the tendency of the algorithm to over-predict residential use, we remove cells zoned as Residential from consideration.  This leaves a nearly equal share of the remaining four uses: Commercial, Industrial, Parks, and Other.  Table \ref{tab:bos_class_result_four}  and Figure \ref{fig:bos_class_result_22} display results for this sub-classifier.  Now, the zone with the largest share is commercial use, which only accounts for 33\% of non-residential zones. Intra-use accuracy has improved significantly for Parks and Other mixed uses.  Whereas the random forest including residential uses could only correctly classify 2\% of zones classified for Parks, the sub-classifier, excluding Residential, correctly predicts 30\% of park cells.  A similar improvement from 10\% to 34\% is also observed for the Other or mixed use category.
The share of classes incorrectly classified as Residential roughly is distributed onto Parks and Others in the classifier without the Residential category, while commercial and industrial zones are not affected heavily. One hypothesis for this effect is that many cells while classified as Residential in rural areas are not fully developed and thus used as parks and in the city center show mixed usage. Including the large class of residential zones masks this effect. 

\begin{table}[h] 
	\caption{Random forest classification results.  In this case, residential land has been removed from consideration.  The algorithm is now able to correctly predict much larger fractions of rarer land uses.}
	\label{tab:bos_class_result_four}
	
\begin{tabular}{cccccc}
\hline 
 Total Accuracy: & \multicolumn{5}{l}{ 0.40} \\
\hline
 & Res & Com & Ind & Prk & Oth   \\
 Land Share:& 0.00 & 0.33 & 0.31 & 0.16 & 0.20\\
 Vote Thresh: & N/A  & 0.30 & 0.30 & 0.20 & 0.20 \\
\hline
& \multicolumn{5}{c}{Confusion Matrix} \\
& Res & Com & Ind &Prk & Oth \\
Res & \cellcolor[gray]{0.8}N/A & N/A  & N/A  & N/A  & N/A   \\
Com & N/A  & \cellcolor[gray]{0.8}0.50 & 0.19 & 0.11 & 0.19 \\
Ind  & N/A  & 0.27 & \cellcolor[gray]{0.8}0.37 & 0.12 & 0.24  \\
Prk &N/A  & 0.31 & 0.18 & \cellcolor[gray]{0.8}0.29 & 0.21 \\
Oth &  N/A  & 0.26 & 0.24 & 0.15 & \cellcolor[gray]{0.8}0.34\\
\hline
		\end{tabular}
	
\end{table}

\begin{figure}[ht] 
\centering
\includegraphics[width=1\linewidth]{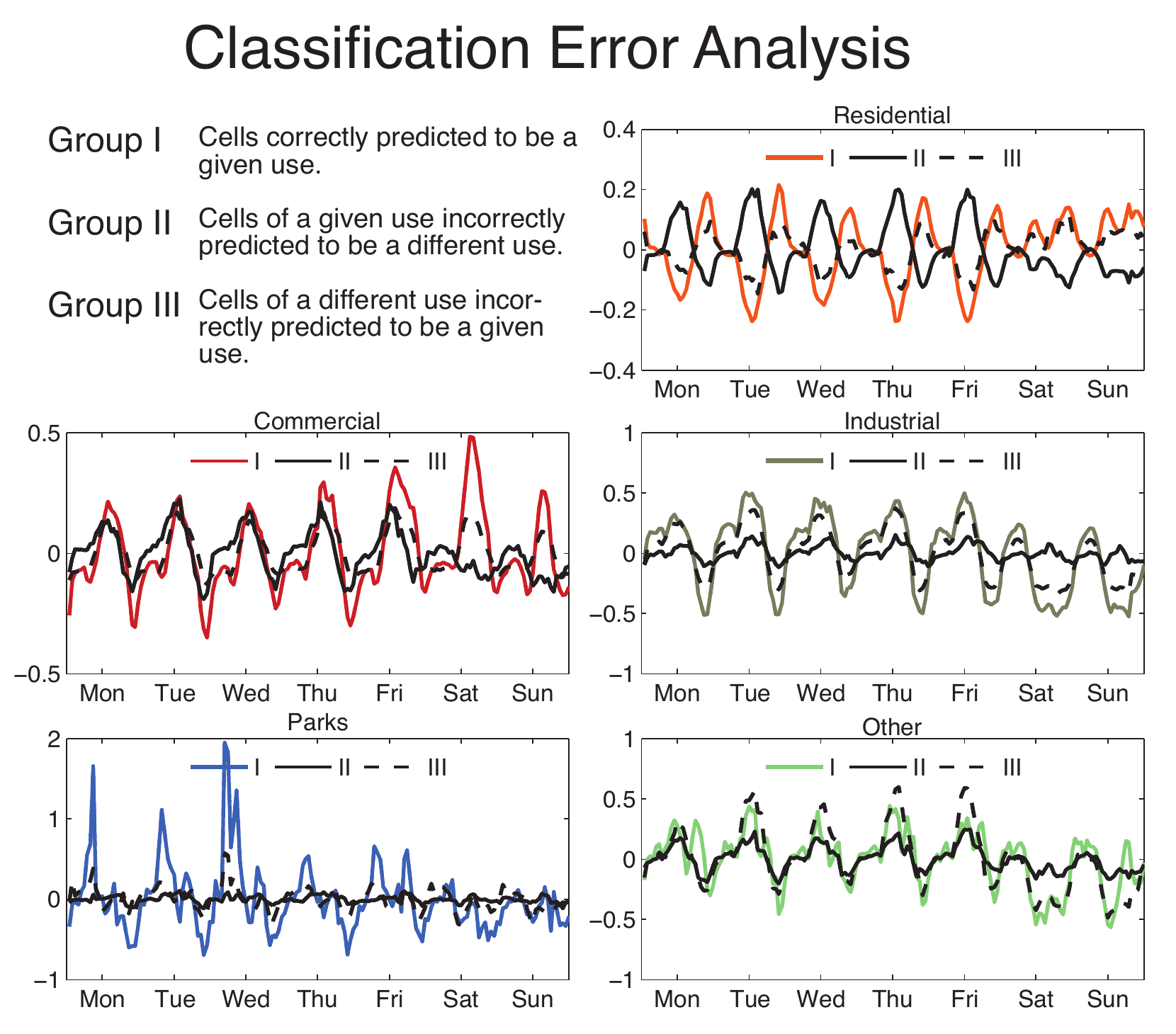}
\caption{An analysis of classification errors. We consider three groups: (I) Cells correctly predicted to be a given use (II) Cells of a given use incorrectly predicted to be some other use (III) Cells of some other use incorrectly predicted to be a given use. For example, Group I includes all residential areas correctly predicted to be residential.  Group II, residential cells predicted to be some other use (i.e. Commercial), have average activity that is the inverse of Group I, suggesting these locations were misclassified because they display fundamentally different activity patterns.  Group III represent cells of other uses such as Commercial that behave like Residential. This error analysis suggests that our algorithm is clustering locations based on both their zoned use as well as the dominant patterns in mobile phone activity.}\label{fig:class_error_ts}
\end{figure}

The goal of the supervised learning algorithm is to make correct predictions of actual zoned use.  Incorrectly classified cells are labeled as errors, but how an area is zoned is not necessarily the same as how it is used. As an example the area termed "Back Bay" containing some of Boston's most busiest shopping streets, Boylston and Newbury, is classified as residential, as is the campus of MIT. Clearly these areas have a different usage than residential areas in the suburbs. A political and idiosyncratic process for setting and updated zoning regulations may lead to broad or unenforced development standards.  In light of this, errors made by our classification algorithm may be due to incomplete zoning data rather than actual mistakes.  To examine this possibility further, we analyze prediction errors more closely. Figure \ref{fig:class_error_ts} displays a detailed partitioning of classifier results. We compare average residual activity across three groups of cells: (I) All cells correctly predicted to be a given use. (II) All cells of another use incorrectly predicted to be the given use. (III) All cells of a given use incorrectly predicted to be some other use.  

Reviewing residential use, we see that Group I is defined as all residential cells correctly predicted to be residential. The average activity pattern is the most dominant pattern of residual activity for residential land use.  We find that the residual activity in non-residential cells predicted to be residential (Group II) closely follows the pattern found in Group I.  This strongly supports our hypothesis that though some zones are not classified as residential in the data, their phone activity patterns suggest they are used in similar ways.  In contrast, the residual activity in residential cells incorrectly classified as some other use (Group III) displays the inverse pattern.  This suggests our algorithm is identifying cells that are zoned as residential use but that do not share activity characteristic of that zoning class in reality.

\section{Conclusion}
In this article, we examined the potential of CDR data to predict land usage. We demonstrated that aggregate data shows the potential to differentiate land usage based on temporal distribution of activities. While the absolute activity is dominated by the circadian rhythm of life, eliminating this rhythm reveals subtle differences between the five main land use categories Residential, Commercial, Industrial, Parks and Other.  The addition of a temporal dimension to zoning classification may aid strategic planning decisions related to land use.

As the data are available at a high spatial resolution, we investigated the capabilities to infer land use on a fine grid of 200 by 200 meters. We found that supervised classification based on labeled zoning data provides estimated land use classifications which show better accuracy than random assignment. At the same time accuracy is worse than classifying every zone as Residential, the dominant category. 

Reasons for this lack of accuracy might be found in the nature of the data used: actual usage might differ from the zoning regulations and Residential is often confused with Parks and Other zones. Omitting residential zones, the classification accuracy for Parks and Other zones greatly increases while industrial and commercial zones classification accuracies are not heavily affected.  For rural areas where residential land might not be fully developed this is plausible. For urban zones the distinction between Residential and Other zones might also be subject to temporal changes as mixed use is prevalent.  Finally, analysis of prediction errors reveals that the algorithm fails to correctly classify areas because they have fundamentally different mobile phone activity patterns.  This suggests that there may be heterogeneity in how land is actually used, despited its official zoned classification.

Thus the main conclusion is that the CDR data shows some potential to infer actual land use both on an aggregate level and on a higher spatial resolution.  However, zoning data might not be the optimal data source to infer actual land use and hence act as ground truth to guide the supervised learning algorithm.  In this respect, our analysis suggests that mobile phone activity may be used to measure the heterogeneity in how space is used that cannot be captured by simple and broad zoning classifications.  Moreover, the incorrect predictions made by our algorithm may suggest updates to traditional zoning maps so as to better reflect actual activity or highlight areas where more planning oversight is needed.

Both topics will be investigated further. Larger sample sizes in the form of longer time series might lead to a reduction in noise levels and hence increase the classification accuracy.  It may also be advantageous to expand the set of features used in prediction.  Although our aim was to keep this space relatively low dimensional to aid interpretation, the complexity of intracity mobility may demand more.  However, given our results suggest that a modest fraction of the city can be classified at very high resolution from relatively simple features.  The algorithm itself may be improved by additional consideration of balancing more prevalent uses with those more scarce. Finally, other data sources such as points of interest (POIs) will be used as ground truth in the supervised learning instead of zoned use.  This may clarify whether the deviations in classification between the zoning regulations and the mobile phone usage dynamics are due to wrong zonings or deficiencies in the measurement technology using CDR data.



We hope this information will be useful to make effective and efficient choices of locations for both public and private resources. In addition to potential applications, we hope that tools and techniques developed and applied above will prove useful to merging traditional and novel data.  

\section{Acknowledgements}
This work was made possible in part by the MIT - Xerox Fellowship as well as an National Science Foundation Graduate Research Fellowship.  In addition, Marta C. Gonz\'{a}lez acknowledges awards from NEC Corporation Fund and the Solomon Buchsbaum Research Fund. The funders had no role in study design, data collection and analysis, decision to publish, or preparation of the manuscript.

\bibliographystyle{abbrv}
\bibliography{jltoole-landuse}  

\begin{thebibliography}{10}

\bibitem{Banister1997}
D.~Banister.
\newblock {Reducing the need to travel}.
\newblock {\em Environment and Planning B: Planning and Design},
  24(3):437--449, 1997.

\bibitem{Breiman2001}
L.~Breiman.
\newblock {Random Forests}.
\newblock {\em Machine Learning}, 45(1):5--32, Oct. 2001.

\bibitem{Calabrese2011}
F.~Calabrese, G.~Di~Lorenzo, L.~Liu, and C.~Ratti.
\newblock {Estimating Origin-Destination Flows Using Mobile Phone Location
  Data}.
\newblock {\em IEEE Pervasive Computing}, 10(4):36--44, Apr. 2011.

\bibitem{Calabrese2010}
F.~Calabrese, J.~Reades, and C.~Ratti.
\newblock {Eigenplaces: Segmenting Space through Digital Signatures}.
\newblock {\em IEEE Pervasive Computing}, 9(1):78--84, Jan. 2010.

\bibitem{Cervero1996}
R.~Cervero and K.~Kockelman.
\newblock {Travel demand and the 3Ds: Density, diversity, and design}.
\newblock {\em Transportation Research Part D: Transport and Environment},
  2(3):199--219, Sept. 1997.

\bibitem{Eagle2006}
N.~Eagle and A.~Pentland.
\newblock {Reality mining: sensing complex social systems}.
\newblock {\em Personal and Ubiquitous Computing}, 10(4):255--268, May 2006.

\bibitem{Geurs2004}
K.~T. Geurs and B.~van Wee.
\newblock {Accessibility evaluation of land-use and transport strategies:
  review and research directions}.
\newblock {\em Journal of Transport Geography}, 12(2):127--140, June 2004.

\bibitem{Gonzalez2008}
M.~C. Gonzalez, C.~A. Hidalgo, and A.-L. Barabasi.
\newblock {Understanding individual human mobility patterns}.
\newblock {\em Nature}, 453(7196):779--782, June 2008.

\bibitem{Jacobs1961}
J.~Jacobs.
\newblock {\em {The death and life of great American cities}}.
\newblock Vintage Books, 1961.

\bibitem{Maat2005}
K.~Maat, B.~van Wee, and D.~Stead.
\newblock {Land use and travel behaviour: expected effects from the perspective
  of utility theory and activity-based theories}.
\newblock {\em Environment and Planning B: Planning and Design}, 32(1):33--46,
  2005.

\bibitem{Reades2009}
J.~Reades, F.~Calabrese, and C.~Ratti.
\newblock {Eigenplaces: analysing cities using the space\"{y} - \"{y}time
  structure of the mobile phone network}.
\newblock {\em Environment and Planning B: Planning and Design},
  36(5):824--836, 2009.

\bibitem{Barabasi2010}
C.~Song, Z.~Qu, N.~Blumm, and A.-L. Barab\'{a}si.
\newblock {Limits of Predictability in Human Mobility}.
\newblock {\em Science}, 327(5968):1018--1021, Feb. 2010.

\bibitem{Soto2011}
V.~Soto and E.~F. Mart'{\i}nez.
\newblock {Automated land use identification using cell-phone records}.
\newblock In {\em Proceedings of the 3rd ACM international workshop on
  MobiArch}, HotPlanet '11, pages 17--22, New York, NY, USA, 2011. ACM.

\bibitem{Yuan2012}
J.~Yuan, Y.~Zheng, and X.~Xing.
\newblock Discovering regions of different functions in a city using human
  mobility and pois.
\newblock {\em KDD}, 2012.

\end{thebibliography}

\balancecolumns
\end{document}